\documentclass[conference]{IEEEtran}
\usepackage{graphicx,times,amsfonts,amssymb,amsmath,multirow,subfigure} 
\usepackage{algorithmic,algorithm,url}

\usepackage{tikz}
\usepackage{textcomp}
\usepackage{hyperref}
\usepackage{lipsum}

\newcommand\copyrighttext{%
  \footnotesize \textcopyright 2017 IEEE. Personal use of this material is permitted.
  Permission from IEEE must be obtained for all other uses, in any current or future
  media, including reprinting/republishing this material for advertising or promotional
  purposes, creating new collective works, for resale or redistribution to servers or
  lists, or reuse of any copyrighted component of this work in other works.\\
  \emph{To appear in the Proceedings of the 2017 International Joint Conference on Neural Networks (IJCNN 2017)}}
\newcommand\copyrightnotice{%
\begin{tikzpicture}[remember picture,overlay]
\node[anchor=north,yshift=-10pt] at (current page.north) {\fbox{\parbox{\dimexpr\textwidth-\fboxsep-\fboxrule\relax}{\copyrighttext}}};
\end{tikzpicture}%
}

\newcommand{\W}{\mathbf{W}}

\newcommand{\Wout}{\W_{out}}

\IEEEoverridecommandlockouts    

\begin{document}


\title{DropIn: Making Reservoir Computing Neural Networks Robust to Missing Inputs by Dropout}
\author{\IEEEauthorblockN{Davide Bacciu and Francesco Crecchi}
\IEEEauthorblockA{Dipartimento di Informatica\\
Universit\`a di Pisa\\
Largo B. Pontecorvo 3\\
Pisa, Italy\\
Email: bacciu@di.unipi.it}
\and
\IEEEauthorblockN{Davide Morelli}
\IEEEauthorblockA{Biobeats Ltd\\
London, UK\\
Email: davide@biobeats.com
}
}


\maketitle
\copyrightnotice

\begin{abstract}
The paper presents a novel, principled approach to train recurrent neural networks from the Reservoir Computing family that are robust to missing part of the input features at prediction time. By building on the ensembling properties of Dropout regularization, we propose a methodology, named DropIn, which efficiently trains a neural model as a committee machine of subnetworks, each capable of predicting with a subset of the original input features. We discuss the application of the DropIn methodology in the context of Reservoir Computing models and targeting applications characterized by input sources that are unreliable or prone to be disconnected, such as in pervasive wireless sensor networks and ambient intelligence. We provide an experimental assessment using real-world data from such application domains, showing how the Dropin methodology allows to maintain predictive performances comparable to those of a model without missing features, even when 20\%-50\% of the inputs are not available.
\end{abstract}

\IEEEpeerreviewmaketitle

\section{Introduction} \label{sect:intro}
The increasing diffusion of networks of pervasively distributed environmental and personal sensor devices requires computational models capable of dealing with continuous streams of sensor data under the form of time series of measurements.
Machine learning models, in this context, serve to make sense of such multivariate sequences of heterogeneous sensor information by providing predictions supporting context-awareness and ambient intelligence functions. Numerous applications have been developed by modeling them as regression and classification tasks on sensor streams, including event recognition, fault and anomaly detection, human activity recognition and, in general, supporting robotic and Intenet-of-Things (IoT) applications by providing adaptivity and context awareness mechanisms.

Recurrent Neural Networks (RNN) are a popular and effective means to deal with sequential information thanks to their ability in encoding the history of the past inputs within the network state. However, these networks, as well as their feedforward counterpart, require a set of fixed input features to be available both at training as well as at a test time. Their predictive performance tends to abruptly decline when one or more of the input features on which it has been trained is missing when querying the model for predictions \cite{krause2003ensemble}. Many practical ubiquitous computing and IoT scenarios comprise information generated by networks of loosely connected devices, which are often battery-operated and communicating through wireless channels with little quality of service guarantees. In such scenarios, it is not unlikely to have to deal with missing information, which might result in a RNN performing recognition or prediction tasks missing a part of its inputs.

The goal of this work is to propose, and to experimentally assess, a principled approach to make neural networks robust to such missing inputs, at query time (i.e. at prediction), focusing in particular on efficient RNN models from the Reservoir Computing (RC) \cite{Lukosevicius2009} paradigm.  The large majority of the works in literature addresses the problem of missing inputs solely with respect to information that is missing at training time. In this context, dealing with missing information amounts to finding the best strategies to impute values of certain features which are missing in some of the training samples: \cite{garcia2010pattern} provides a recent survey on this problem. In this work, instead, we consider a scenario in which we are able to train a RNN using complete data (i.e. with no missing inputs) but where part of the inputs may become unavailable, even for a long time, during system operation (e.g. due to a sensor failure). One of the typical approaches to deal with this problem exploits imputation, i.e. substituting the missing feature by inferring it from other information. In \cite{parkerArt}, for instance, a spatial-temporal replacement scheme is proposed for a fuzzy Adaptive Resonance Theory (ART) neural network used for anomaly detection over a Wireless Sensor Network (WSN). Here missing input information related to a WSN node is imputed based on a majority voting between the readings of the devices that are spatially closer to the failing one. Similarly, \cite{parkerKnn} uses a k-Nearest Neighbor (kNN) to replace missing sensor values with spatially and temporally correlated sensor values. This approach has a major drawback in the fact that it assumes that sensor devices are of homogenous type and that it is possible to find a spatially/temporally related sensor, of the same type as the failing one, to replace its measurements. An alternative approach to the problem is based on fitting a probability distribution of the missing features given the observable inputs, e.g. by using a Parzen window estimator \cite{tresp1995efficient}, and to use it to sample replacement measurements for the missing information \cite{tresp1994training,tresp1995missing}. In \cite{bengio1996recurrent}, it is proposed a discriminative solution that uses a RNN to model static (i.e. non sequential) problems while recurrent layers are used to estimate the values of the missing features. A rather different approach to information imputation uses committee machines or model ensembles \cite{krause2003ensemble}: these models accommodate data with missing features by training an ensemble of predictors with random subsets of the total number of available features.

None of the approaches described above attempt to make the original model robust to missing information. They either impute the missing information or enrich the original model by training multiple instances of it with different input configurations, with considerable impact on the training time as the number of input features increases. We propose a novel solution to the missing input problem that exploits \emph{dropout} \cite{dropout2014}, a regularization technique that has become widely popular in the context of deep learning in the latter years. Dropout prevents overfitting by randomly selecting, during training, a subset of neurons or synapses which are dropped (i.e. their activation is zeroed) for the current training sample. The key intuition guiding this work is that one of the effects of Dropout is to train the full neural model as if it were an ensemble of thinned models obtained by the random dropout disconnections. As such, we believe that it can be exploited to build a model robust to missing inputs as an ensemble of thinned models obtained by applying dropout to the input neurons and connections alone.

In the following, we discuss an application of such a technique to Reservoir Computing (RC) \cite{Lukosevicius2009}, a family of efficient RNN models which is quite popular in applications dealing with sensor streams due to its robustness to noise and computational efficiency, e.g. see \cite{bacciu2013experimental} for an embedded implementation on low-power WSN devices. RC networks are based on the separation between a recurrent dynamical layer, i.e. the reservoir, and a non-recurrent output layer, i.e. the readout. In particular, we will focus on the Echo State Network (ESN) model \cite{Jaeger2004} where only the output layer connections are trained while the input and recurrent weights are left randomly initialized. Note that this is the first application of the dropout techniques to RC models and it is also the first attempt to make these networks robust to missing inputs. The choice of ESN as reference model is motivated, on the one hand, by its relevance and popularity in ambient intelligence, pervasive computing and IoT applications. On the other hand, we would like to assess the effect of our dropout approach, referred to as \emph{DropIn} from here onwards, on the most challenging conditions, i.e. where only the output connections are adaptive to see if the beneficial effect of the input dropout propagates to them. Nevertheless, the proposed DropIn technique can be applied to any recurrent or feedforward neural network, as in the standard dropout technique.

The remainder of the paper is organized as follows: Section \ref{sect:back} provides a brief background on dropout regularization and on the reservoir computing paradigm; Section \ref{sect:approach} describes DropIn application to ESN, while Section \ref{sect:expcomp} provides an experimental assessment of the proposed model on two real-world datasets from ambient intelligence and pervasive computing applications. Section \ref{sect:conclude} concludes the paper.

\section{Background} \label{sect:back}
In this section, we provide a brief overview of the Dropout approach, in Section \ref{sect:drop}, followed by an introduction to RC and ESN models in Section \ref{sect:rc}.

\subsection{Dropout Regularization} \label{sect:drop}
Dropout is a regularization technique for fully connected neural network layers with adaptive synaptic weights \cite{dropout2014}. The key intuition of Dropout is to prevent units from co-adapting by randomly disconnecting a subset of them from the network during training. Dropout is also though to act as a model combination, or ensembling technique such that it can efficiently combine an exponential number of neural network architectures corresponding to the thinned networks obtained by randomly dropping the neuron subsets.

Dropout training works by randomly determining which units are kept at a given training instant; each unit is retained with a fixed probability $p$, independently from other units, or is dropped with probability $1-p$. The retention probability $p$ is typically determined by validation. A dropped out unit is temporarily removed from the network, along with all its incoming and outgoing connections: in practice, this amounts to zeroing the dropped unit output. In this respect, a neural network with $N$ units is an ensemble of $2^N$ thinned networks whose total number of parameters is still bound by $O(N^2)$ since they all share the weights. At test time, it is not feasible to reproduce the network thinning process to compute the ensemble prediction from all the $2^N$ models. However, \cite{dropout2014} shows how this can be efficiently approximated by using the original un-thinned network with weights scaled by the retention probability $p$.

The Dropout approach has been later generalized by the DropConnect model \cite{wan2013regularization}. Instead of dropping out single units, DropConnect sets a random subsets of the network weights to zero, such that each unit receives input from a portion of neurons in the previous layer. To do so, it exploits, again, a retention probability $p$ which is independently applied to the single elements of the neural network connectivity matrix. This is shown to enhance the ensembling effect with respect to Dropout, as the number of thinned network in DropConnect is $O(2^{|M|})$, where $M$ is the matrix of randomly selected zero and ones used for weight masking \cite{wan2013regularization}.

\subsection{Reservoir Computing} \label{sect:rc}
Reservoir Computing \cite{Lukosevicius2009} is a RNN paradigm based on the separation between the recurrent part of the network, the \emph{reservoir}, from the feedforward neurons, including the output layer referred to as  \emph{readout}.
The reservoir encodes the history of the input signals. The activations of its neurons (i.e. the network state) are combined by the readout layer to compute the network predictions. One of the key aspect of RC is its training efficiency as the readout layer is the only trained part of the network, whereas the input and reservoir connections are randomly initialized, under conditions ensuring contractivity \cite{JaegerTech2001}, and then left untrained.

Among the different RC approaches, here we focus on the popular ESN model \cite{Jaeger2004} due to its suitability to the ambient intelligence applications considered in this work. An ESN comprises an input layer with $N_U$ units, a reservoir layer with $N_R$ units and a readout with $N_Y$ units.  The reservoir is a large, sparsely-connected layer of recurrent non-linear units (typically $tanh$) which is used to perform a contractive encoding \cite{JaegerTech2001} of the history of driving input signals into a state space. The readout comprises feedforward linear neurons computing ESN predictions as a weighted linear combination of the reservoir activations. In this work, we focus on the leaky integrator ESN (LI-ESN), a variant of the standard ESN model which applies an exponential moving average to the reservoir state space values. This allows a better handling of input sequences that change slowly with respect to sampling frequency \cite{Lukosevicius2009} and it has been shown to work best, in practice, when dealing with sensor data streams \cite{bacciu2013experimental}.
At each time step $t$, the activation the reservoir activation of a LI-ESN is computed by
\begin{equation}
\label{eq.reservoir}
\mathbf{x}(t) = (1 - a) \mathbf{x}(t - 1) + a f(\mathbf{W}_{in} \mathbf{u}(t) + \mathbf{W}_h \mathbf{x}(t - 1))
\end{equation}
where $\mathbf{u}(t)$ is the vector of $N_U$ inputs at time $t$, $\mathbf{W}_{in}$ is the $N_R \times N_U$ input-to-reservoir weight matrix, $\mathbf{W}_h$ is the $N_R \times N_R$ recurrent reservoir weight matrix and $f$ is the component-wise reservoir activation function.  The term $a \in [0,1]$ is a \emph{leaking rate} which controls the speed of LI-ESN state dynamics, with larger values denoting faster dynamics.

Reservoir parameters are left untrained after a random initialization subject to the constraints given by the so called \emph{Echo State Property} (ESP) \cite{JaegerTech2001}, requiring that network state asymptotically depends on the driving input signal and any dependency on initial conditions is progressively lost. In  \cite{JaegerTech2001}, it is provided a necessary and a sufficient condition for the ESP. The sufficient condition states that the largest singular value of the reservoir weight matrix must be less than $1$. For a LI-ESN model, such condition applies to the leaky integrated matrix
\[
    \tilde{\mathbf{W}} = (1-a)\mathbf{I} + a \mathbf{W}_h,
\]
such that
\begin{equation}
\label{eq.suff}
    \sigma(\tilde{\mathbf{W}}) < 1
\end{equation}
where $\sigma(\tilde{\mathbf{W}})$ is the largest singular value of $\tilde{\mathbf{W}}$. The necessary condition \cite{JaegerTech2001}, on the other hand, says that if the \emph{spectral radius} $\rho(\tilde{\mathbf{W}})$, i.e. the largest absolute eigenvalue of the matrix $\tilde{\mathbf{W}}$, is larger than $1$, the network as an asymptotically unstable null states and hence lacks the ESP. The sufficient condition in (\ref{eq.suff}) is considered by to be too restrictive \cite{JaegerTech2001} for practical purposes. Instead, the $\tilde{\mathbf{W}}$ matrix is often initialized to satisfy the necessary condition, i.e.
\begin{equation}
\label{eq.initialization}
\rho(\tilde{\mathbf{W}}) < 1,
\end{equation}
with values of the spectral radius that are, typically, close to the stability threshold $1$. Input weights are randomly chosen from a uniform distribution over $[-s_{in}, s_{in}]$ (where $s_{in}$ is an input scaling parameter), while $\mathbf{W}_h$ is typically from a uniform distribution in $[-1,1]$ and then scaled so that Eq.~\ref{eq.initialization} holds.

The LI-ESN output is computed by the readout through the linear combination
\begin{equation}
\label{eq.readout}
\mathbf{y}(t) = \mathbf{W}_{out} \mathbf{x}(t)
\end{equation}
where $\mathbf{W}_{out}$ is the ${N_Y \times N_R}$ reservoir-to-readout weight matrix. The readout output can be computed for each time step $t$ or only for a subset of them depending on the application (e.g. for sequence classification only the output corresponding to the last sequence item is computed, typically). Training of an ESN model amounts to learning the values of the $\Wout$ matrix which implies the solution of a linear least squares minimization problem: this is typically achieved by  efficient batch linear methods such as Moore-Penrose pseudo-inversion and ridge regression \cite{Lukosevicius2009}.

\section{DropIn Reservoir Computing} \label{sect:approach}

We describe a novel use of the Dropout technique as a principled approach to make neural networks robust to missing inputs at prediction time. We name this approach DropIn as it is based on the use of unit-wise Dropout \cite{dropout2014} at the level of the input neurons of the network. The rationale inspiring our work is the observation that the application of Dropout to $N_U$ input units is essentially training a committee machine of $2^{N_U}$ thinned networks with shared weights. In this sense, our approach recalls the committee machine by \cite{krause2003ensemble}. Differently from this work, our method does not require to define a specific committee machine architecture nor it induces any increase in the parametrization of the original model. The choice of using unit-wise Dropout \cite{dropout2014} in place of the connection-wise DropConnect \cite{wan2013regularization} is motivated by the fact that the latter approach would not make the neural network robust to losing one or more of its inputs; rather, it would make the network somewhat less focused on the exploitation of a specific input which, in principle, can be very discriminant for the task. Some preliminary experiments performed with the DropConnect approach (not reported in this paper for brevity) confirm our intuition that it performs significantly worse than Dropout on our tasks.

The DropIn approach is general, as it can be applied to any neural network model for which Dropout applies. Here, we focus on the use of DropIn in the context of RC and, in particular, with the ESN model. The motivation for such a choice is twofold. First, RC models are a popular approach in ambient intelligence, pervasive computing and IoT applications \cite{bacciu2013experimental,Dragone2015269,Obst2014}, that are  areas where one has to deal with input information collected by low-fidelity devices (e.g. battery operated) and transmitted over loose communication channels, thus involving an high likelihood of faults and missing data \cite{parkerKnn}. The second motivation is more related to the exploration of the effect of DropIn in a learning paradigm which is, perhaps, the least straightforward in terms of Dropout application. The characteristics of RC, in fact, makes it benefit poorly from Dropout in terms of regularization, as only a minor part of the network weights are trained. In fact, to the extent of our knowledge, this is the first work in which Dropout techniques are being used in RC. Here, we seek to determine if the application of Dropout to inputs with untrained weights has an effect propagating throughout the untrained reservoir all the way to the readout, which is the only trained part of the network and hence the only part where the effect of DropIn can be \emph{recorded}.

The application of DropIn to an ESN requires it to be trained through an iterative learning algorithm. In fact, in order to make the network robust to missing inputs, we need the ESN to process the single training sequences multiple times, each time with a fixed probability $1-p$ of independently missing the single inputs, where $p$ is a retention probability as in Dropout.
As discusses in Section \ref{sect:rc}, training an ESN entails solving a least squares minimization problem to find the readout weights $\mathbf{W}_{out}$. The simplest approach would tackle this problem as a gradient descent in the direction of minimizing the instantaneous squared error (at time $t$) $\|\mathbf{y}^{*}(t) - \mathbf{y}(t)\|^2$, where $y^{*}(t)$ is the desired prediction at time $t$. Alternatively, the Recursive Least  Squares (RLS) algorithm has been proposed in \cite{Jaeger2004} as a fast online learning approach. At each time step, RLS minimizes the discounted error
\begin{equation}\label{eq:rls}
E(\mathbf{y},\mathbf{y}^{*}) = \frac{1}{N_Y} \sum_{t=1}^{T} \lambda^{T-t} \|\mathbf{y}^{*}(t) - \mathbf{y}(t)\|^2
\end{equation}
where $0 \leq \lambda \leq 1$ is an error forgetting factor.

\begin{figure}[tb]
  \centering
  \includegraphics[width=\columnwidth]{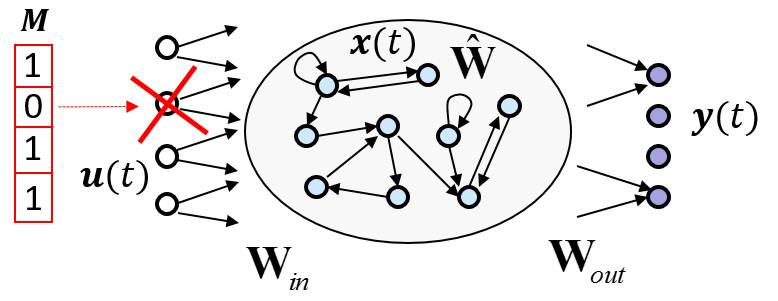}
  \caption{ESN model with DropIn application: the input masking vector $\mathbf{M}$ is generated each time a training sequence is fed to the network. The decision to retain an input unit (i.e. value $1$ in vector $\mathbf{M}$) is taken independently for each feature by random sampling from a uniform probability with retention probability $p$.}\label{fig:esn}
\end{figure}
Our DropIn-ESN model embeds input units dropout within the steps of incremental RLS learning. A schematic view of the procedure is provided by the pseudo-code in Algorithm \ref{alg:dropin} for the ESN architecture depicted in Fig. \ref{fig:esn}. Assuming we are provided with a properly initialized ESN (see Section \ref{sect:rc}), the procedure runs for several epochs through the dataset (each time reshuffling the order of sequences) until an error stability criterion is met (e.g. early stopping on validation error) or the maximum number of epochs is reached. The DropIn mask $\mathbf{M}$ in Fig. \ref{fig:esn} is computed before processing a whole sequence, hence the mask is the same for all the elements of the sequence (see the first operation in the outer for loop in Algorithm \ref{alg:dropin}). This ensures that the readout weights are modified to take into account reservoir activations corresponding to certain inputs being missing. Clearly, the same training sequence can be expected to be processed again in the next epochs, but this time with different input features missing. As in standard Dropout, the decision to drop an input unit is taken independently for each input feature with probability $(1-p)$. Input masking can then be easily obtained by zeroing the elements of current input $\mathbf{u}_i(t)$ using the masking vector $\mathbf{M}$. The equations in the innermost loop of Algorithm \ref{alg:dropin} implement the standard RLS algorithm for ESN, whose details can be found in \cite{Jaeger2004}. The DropIn-ESN model in Algorithm \ref{alg:dropin} has been implemented in Matlab \cite{matlab}: the reference source code used for the experimental assessment in Section \ref{sect:expcomp} can be found here\footnote{\url{https://github.com/FrancescoCrecchi/DropIn-ESN}}.
\begin{algorithm}[tb]
\caption{DropIn Training of an ESN by RLS}
\label{alg:dropin}
\begin{algorithmic}
 \REQUIRE A dataset of $L$ pairs of input-output sequences $(\mathbf{u}_1,\mathbf{y}_1^*),\dots,(\mathbf{u}_L,\mathbf{y}_L^*)$; a properly initialized ESN model; a forgetting factor $\lambda$; a regularization term $\delta$ and an input unit retention probability $p$.
 \STATE Init $\mathbf{S}^{-1}(0) = \delta^{-1} \cdot \mathbf{I}$
 \STATE $n=0$
 \WHILE{learning not converged}
     \STATE Shuffle sequence order
     \FOR{$i = 1$ to $L$}
        \STATE Compute masked inputs vector $\mathbf{M}$ using $p$
         \FOR{$t = 1 \ \mathbf{to} \ T_j$}
            \STATE $n=n+1$
            \STATE Mask current inputs $\mathbf{u}_i(t)$ using $\mathbf{M}$
            \STATE Compute reservoir activation using (\ref{eq.reservoir})
            \STATE Compute readout activation using (\ref{eq.readout})
            \STATE Compute error $\mathbf{e}(n) = (\mathbf{y}_i^*(t) - \mathbf{y}_i(t))$
            \STATE $\mathbf{\Phi}(n) = S^{-1}(n-1) \cdot \mathbf{x}_i(t)$
            \STATE $\mathbf{K}(n) = \mathbf{\Phi}(n) \cdot (\lambda + \mathbf{\Phi}(n) \cdot \mathbf{x}_i(t))^{-1}$
            \STATE $\mathbf{S}^{-1}(n) = \lambda^{-1} (\mathbf{S}^{-1}(n-1) - \mathbf{K}(n) \cdot \mathbf{\Phi}(n))$
            \STATE $\mathbf{W}_{out}(n) = \mathbf{W}_{out}(n-1) + (\mathbf{K}(n) \cdot e(n))^T $
         \ENDFOR
     \ENDFOR
 \ENDWHILE
 \RETURN $\mathbf{W}_{out}$
\end{algorithmic}
\end{algorithm}

Once an ESN model has been trained using DropIn, it can be used for prediction \emph{without} any weight re-scaling. Note that this is different from how standard Dropout works. In a DropIn-ESN, in fact, masking of the input units does not bear any effect on the input-to-reservoir weights, whose values are frozen to their random initialization. Readout weights are the only ones that are affected by DropIn but the effect on them is indirect and mediated by the reservoir activations. Preliminary experiments on the effect of weight re-scaling (not reported here) show that it negatively affects predictive performance.
The reasons underlying such effect deserve further studies: possibly, the fact that the DropIn networks do not require re-scaling might be associated with the fact that at testing time we are simulating missing inputs, hence allowing the network to work under conditions similar to those of the dropout training phase.

\section{Experimental Results} \label{sect:expcomp}
We provide an experimental assessment of the effect of DropIn in two real-world ambient intelligence applications collected within the scope of the European project RUBICON \cite{Dragone2015269}. Both applications involve a WSN, for which we simulate the effect of sensor devices faults at prediction time. For the sake of this work, we assume instead that a sufficient amount of training data without missing information is available, as tackling with missing information at training is outside of the scope of this work. Section \ref{sect:setup} briefly describes the two case studies and summarizes the experimental setup, while Section \ref{sect:results} analyzes the results of the experiments.

\subsection{Data and Experimental Setup} \label{sect:setup}
The first dataset is a human movement prediction benchmark based on WSN information which has been originally presented in \cite{bacciu2013experimental} and that is available for download on the UCI repository\footnote{\url{https://archive.ics.uci.edu/ml/datasets/Indoor+User+Movement+Prediction+from+RSS+data}}. This task deals with the prediction of room changes in an office environment, based on the Received Signal Strength (RSS) of packets exchanged between a WSN mote worn by the user and four anchor devices deployed at fixed positions in the corners of the rooms. The experimental data in \cite{bacciu2013experimental} consists of different setups: in this work, we refer to the \emph{homogenous} setup described in the original paper. Such data is composed of $210$ RSS sequences concerning trajectories of different users moving between $2$ room couples separated by an hallway. The trajectories start at different corners of the rooms and are such that some of them lead to the user changing room while others end up with the user remaining in the same room. The task requires the learning model to predict whether the user is going to change room or to stay in the current one based on RSS information recorded from the starting point of the trajectory. The prediction concerning room change has to be taken at a marker point P, located at $0.6 m$ from the door, which is the same for all the movements; therefore, different paths cannot be distinguished based only on the RSS values collected at M.  This benchmark is formalized as a binary classification task on the RSS time series.  The collected measurements denote RSS samples (integer values ranging from 0 to 100) gathered by sending a beacon packet from the anchors to the mobile at regular intervals, 8 times per second, using the full transmission power of an IRIS-type mote. The RSS values from the four anchors are the input features for the model and are organized into sequences of varying length corresponding to trajectory measurements from the starting point until marker P. A target classification label is associated to each input sequence to indicate whether the user is about to change room (label $+1$) or not (label $-1$). This classification label is provided only for the last element of each training sequence and it thus predicted only for the last time step of each test sequence.

The second dataset, referred to as Kitchen Cleaning \cite{Bacciu2014}, concerns an Ambient Assisted Living scenario where a cleaning robot operates in a home environment located in the {\"A}ngen senior residence facilities in {\"O}rebro Universitet. In this task, the learning model is required to learn to predict a user preference that was not modelled in the robotic planner domain knowledge, that is the fact of not having the robot cleaning the kitchen when the user is in. This task is part of a larger effort to assess the self-adaptation abilities of a robotic ecology planner developed as part of the RUBICON project, including also automated feature selection mechanisms on time series \cite{Bacciu2016}. The experimental scenario consists of a real-world flat sensorized by an RFID floor, a mobile robot with range-finder localization and a WSN with six mote-class devices. Each device is equipped with light, temperature, humidity  and passive infrared (PIR) presence sensors. For our Kitchen Cleaning experiment, we consider only those inputs which have been deemed relevant by the feature selection analysis in \cite{Bacciu2016}, that are the PIR readings of the first five motes plus the information on the $x$ position of the robot in the environment as measured by the range-finder localization system. The dataset contains information on the robot initiating $104$ Kitchen Cleaning tasks: these consist in the robot moving from its base station in the living room to the kitchen, where a user might be present or not or might enter the kitchen during robot navigation. Information is recorded (sampling at $2$Hz) from the WSN and the robot localization system across the whole robot navigation towards the kitchen, thus collecting $104$ time series of $6$ input features. One half of them is associated to a cleaning task that is correctly completed, while the remainder are associated to task failures due to the presence of the user in the kitchen. The target is to learn the Kitchen Cleaning task preference, that is a target output equal to $0$ when the user enters the kitchen at any point of the cleaning task, while this is set to $1$ when the kitchen is free. Differently from the previous dataset, target values are associated to each element of the input sequence and a prediction is thus performed at each time step $t$.

The experimental assessment is intended to confront the performance of a standard LI-ESN, i.e. trained without dropout, with respect to a DropIN LI-ESN model, using different input unit retention probabilities $p \in \{0.8, 0.5, 0.3\}$. The latter $p$ value, in particular, has been added to assess the performance break-point of introducing input dropout.  A model selection scheme has been set-up to assess learning performance. First, for each dataset, a hold-out set of $20\%$ of the total sequences has been extracted to create an external test set. A k-fold cross-validation approach is applied to the remaining $80\%$ of the data for model selection purposes, with $k=5$ and $k=3$ for the first and second dataset, respectively.
Both standard LI-ESN and DropIn LI-ESN have been trained by RLS with hyperparameters varying as follows: number of reservoir neurons in $[50, 100, 300, 500]$ with $10\%$ connectivity, leaky parameter $\alpha \in [0.1, 0.2, 0.3, 0.5, 1]$ and RLS term $\delta \in [0.001, 0.01, 0.1, 1, 10, 100, 1000]$. The $\lambda$ value for RLS has been fixed to $0.9999995$ as discussed in \cite{Jaeger2004}. For each configuration of the hyperparameters, we have generated three random reservoir topologies and random weight initializations using a uniform distribution in $[-0.4,0.4]$ for both input and reservoir weights, along the lines of \cite{bacciu2013experimental}. The reservoir weights have been re-scaled to satisfy the necessary ESP condition, such that the spectral radius $\rho(\tilde{\mathbf{W}}) = 0.99$. A validation error is computed for each hyperparameter configuration (by averaging over the $k$-folds and over the three network topologies) to select the best model: this is then trained on the full training data and tested on the hold-out information.
\begin{table*}[tb]
   \renewcommand{\arraystretch}{1.3}
   \caption{Model selection and testing results for the movement prediction task. The table reports information on the selected configuration (number of reservoir neurons $N_R$, leaky parameter $\alpha$ and RLS term $\delta$) as well as the average classification accuracy in training and validation over the multiple random initializations (standard deviation is in brackets). Test accuracy is reported for the best configuration selected in validation.}\label{tab:aal}
   \centering
  \begin{tabular}{|l|c|c|c|c|c|c|}
  \hline
  Model & $N_R$ & $\alpha$ & $\delta$ & Training & Valid & Test\\
  \hline
  LI-ESN  & 500 &	0.1	& 0.1 & 1.000 (0) & 0.9326 (0.025) & 0.9206\\
  DropInESN (0.8)	& 500 &	0.2	&  0.001 &	0.9494 (0.027) & 0.867 (0.055) & 0.8333\\
  DropInESN (0.5)	& 50   & 0.1 & 0.1 & 0.7386 (0.032) & 0.6786 (0.068) & 0.7778\\
  DropInESN (0.3)	& 500  & 0.3& 0.1 & 0.6052 (0.029) & 0.5734 (0.056) & 0.5714\\
  \hline
 \end{tabular}
\end{table*}

\subsection{Results and Analysis} \label{sect:results}
The aim of this experimental assessment is to assess the effect of missing inputs on the predictive performance of the LI-ESN trained with and without DropIn and using different unit retention probabilities. To this end, we first provide baseline results for the models without missing inputs. Then, we show the effect of an increasing number of missing input features, by performing test predictions with input features removed. For instance, the performance with a single missing input is obtained by removing a single input feature for each test sequence at a time, then the network prediction is computed and the process is iterated for all the input features and an average prediction performance is computed. Similarly, when testing multiple missing inputs, we perform predictions using all possible combinations of missing features and we then provide performance statistics averaged on all combinations.

Table \ref{tab:aal} shows the baseline results (no missing inputs) and the model selected configuration for the first dataset. Here model performance is assessed by classification accuracy. These baseline results show that the LI-ESN model trained without DropIn achieves higher validation and test accuracies, while decreasing levels of input retention probabilities yield to decreasing predictive accuracies when all inputs are present. This is not surprising as, for instance, a retention probability $p=0.3$ on this task means, essentially, that the model is being trained to perform predictions using only information from a single input randomly selected each time from the available four. When using $p=0.8$, the DropIn ESN performance is not far from that of the standard LI-ESN: the performance difference between the two can be explained in the terms of the excessive regularization the DropIn-ESN is subject to, which yields to an underfitting. This can be clearly seen by confronting the training accuracy of the models, where standard LI-ESN achieves an $100\%$ classification accuracy, while the DropIn-ESN (with $p=0.8$) stops at $91.46\%$.

\begin{figure}[tb]
  \centering
  \includegraphics[width=\columnwidth]{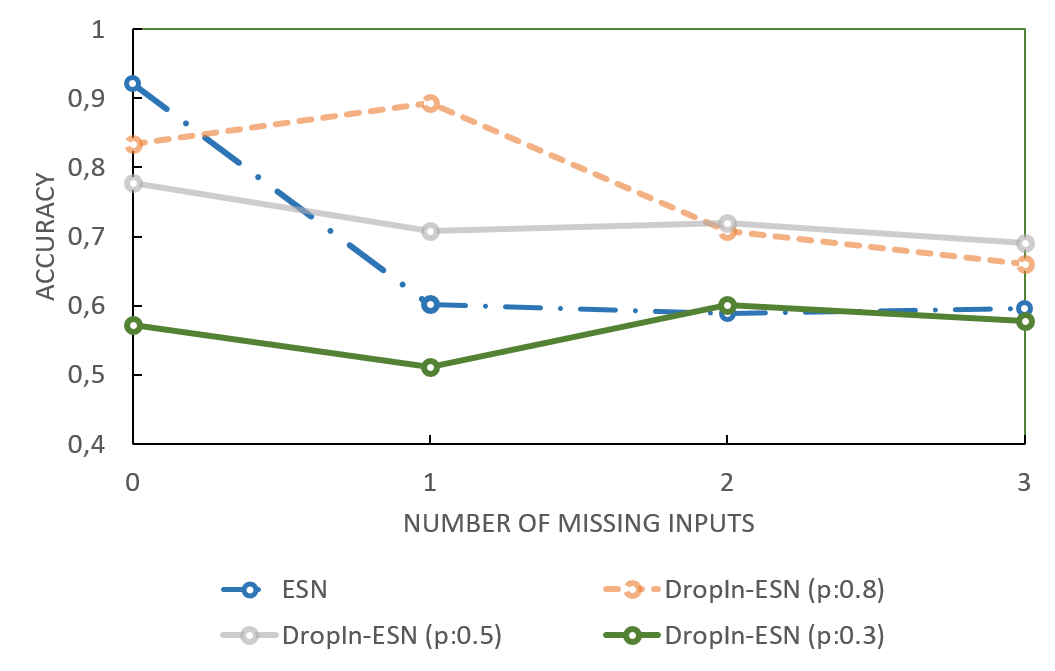}
  \caption{Test accuracy of the LI-ESN vs DropInESN as a function of the number of missing inputs for the movement prediction task.}\label{fig:aal}
\end{figure}

Figure \ref{fig:aal}, on the other hand, shows clearly the advantage of DropIn training when dealing with missing inputs. The predictive performance of a standard LI-ESN drops considerably (to about $60\%$) already when a single input is missing. On the contrary, the performance of the DropIn network with $p=0.8$ increases with respect to the baseline reaching $90\%$ and remaining well above the standard LI-ESN model even when more inputs are lost. The reason for the performance increase with respect to baseline has to be sought in the fact that by dropping a single input at test time we are basically allowing the network with $p=0.8$ to work under the same training conditions, i.e. missing on average one input each time. The performance of DropIn when using  $p=0.5$ is also better than LI-ESN, whereas $p=0.3$ is clearly the breakpoint for the amount of inputs that can be dropped in training for this task.

Table \ref{tab:kitchen} shows the baseline results for the Kitchen Cleaning data: here performance is assessed using Mean Absolute Error (MAE) as targets and outputs are available for each element of the sequence. On this dataset, the beneficial effect of the DropIn training is even more evident. Its strong regularization effect, in this case, ensures that all DropIn networks yield to better test set performances than standard LI-ESN even when using all the features, despite the fact that the latest model achieves the lowest validation error (also the lowest training error, not shown here). The DropIn network with $p=0.3$ is the one that fits less the training data and, yet, it generalizes better to the test set. This might be due to the specifics of the task which, in particular, is characterized by the presence of $5$ PIR inputs which have an high degree of redundancy with each other, hence favouring the more constrained model.
\begin{table*}[tb]
   \renewcommand{\arraystretch}{1.3}
   \caption{Model selection and testing results for the Kitchen Cleaning task. The table reports information on the selected configuration (number of reservoir neurons $N_R$, leaky parameter $\alpha$ and RLS term $\delta$) as well as the average MAE in training and validation over the multiple random initializations (standard deviation is in brackets). Test error is reported for the best configuration selected in validation.}\label{tab:kitchen}
   \centering
  \begin{tabular}{|l|c|c|c|c|c|c|}
  \hline
  Model & $N_R$ & $\alpha$ & $\delta$ & Training & Valid & Test\\
  \hline
  LI-ESN  & 500 &	0.3	& 10 & 0.0655 (0.033) & 0.1831 (0.058) & 0.2136\\
  DropInESN (0.8)	& 500 &	0.5	& 10 & 0.0904 (0.029) & 0.2056 (0.092) & 0.1764\\
  DropInESN (0.5)	& 100   & 0.5 & 0.1 & 0.1829 (0.015) & 0.2182 (0.055) & 0.1810\\
  DropInESN (0.3)	& 500  & 0.5 & 100 & 0.1834 (0.037) & 0.2378 (0.063) & 0.1471\\
  \hline
 \end{tabular}
\end{table*}

\begin{figure}[tb]
  \centering
  \includegraphics[width=\columnwidth]{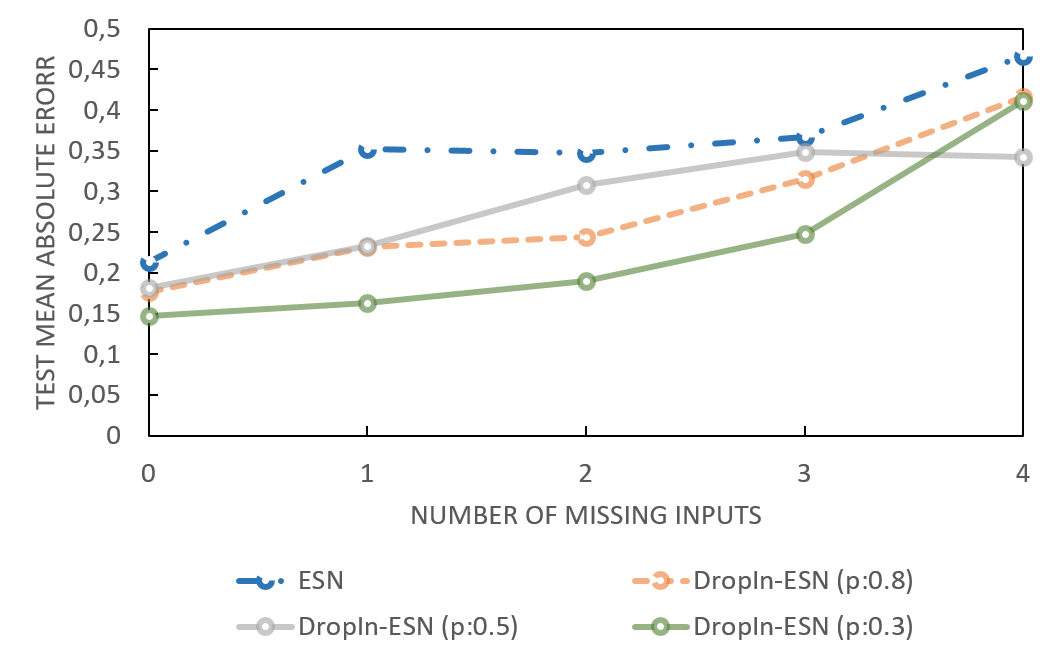}
  \caption{Test MAE of the LI-ESN vs DropInESN as a function of the number of missing inputs for the Kitchen Cleaning task: a maximum of $4$ PIR inputs is assumed to be missing altogether.}\label{fig:kitchen}
\end{figure}

The advantage of the DropIn approach is evident especially when considering the effect of missing inputs, as shown by the plot in Fig. \ref{fig:kitchen}. Again, the standard LI-ESN model does not cope well already with a single missing input feature, with a MAE jumping to $0.35$ and raising up to about $0.45$ when $4$ inputs are missing. On the other hand, the DropIn model with $p=0.3$ has the best MAE for most of the missing inputs configurations: in particular, the loss of a single input does not shift significantly its error with respect to the baseline in Table \ref{tab:kitchen}.

Summarizing, the results of this preliminary analysis suggest that the DropIn approach can indeed be effective in dealing with missing inputs at prediction time, although this might come at the cost of a slightly reduced performance when all inputs are available, depending on the characteristics of the task. In terms of lesson learned it seems reasonable, when deploying a predictor which can be subject to missing inputs, to train two models: the first is a standard predictor trained without DropIn to work at maximum predictive performance when all inputs are available. The second is a DropIn trained model to be queried when one or more of the inputs is missing to maintain the highest achievable predictive performances. This seems a feasible and practical approach, given the low computational impact of the DropIn procedure. Nevertheless, as suggested by the performance on the second dataset, a DropIn trained model can also be a good candidate for being the sole predictor being deployed for some specific tasks.

\section{Conclusions} \label{sect:conclude}
We have proposed a novel approach to deal with the problem of missing input information at prediction time in neural networks, which are related to the well known Dropout regularization. Our approach, named DropIn, applies Dropout at the level of input neurons. We exploit the ensembling effect produced by Dropout to train a neural network that can be interpreted as a committee of (potentially exponential) sub-networks, each capable of making predictions using only a subset of the available inputs as determined by the neuron retention probability. The proposed approach is principled and general, as it can be seamlessly applied to any artificial neural network model for which Dropout applies and it does not require to define ad-hoc committee architectures or to define data imputation algorithms and models. Further, DropIn is simple to embed in the training phase of the neural model and has a minor computational impact.

We have assessed DropIn performance in conjunction with Reservoir Computing neural models. These are the least straightforward in terms of dropout application due to the untrained nature of their input and recurrent connection weights, which makes recording of the DropIn effect in the synaptic connections quite challenging. At the same time, to the extent of our knowledge, this is the first time in which Dropout techniques are applied to RC models as well as it is the first work specifically addressing the problem of missing input at prediction time in RNNs.

The experimental assessment provides a snapshot of the potential of the approach when applied to real-world predictive tasks comprising input information collected by networks of distributed, fragile and loosely coupled sensor devices. In particular, we have shown how ESN models trained with DropIn can maintain a good predictive performance even when a pair of input features is missing, e.g. due to a sensor fault. Conversely, an ESN model trained with the standard procedure has a neat predictive performance degradation already when missing a single input. Such experimental outcome suggests that DropIn can become a useful methodology when developing ambient intelligence, IoT and pervasive computing applications that need to continuously stream their predictions despite some of the inputs being missing due to device or communication faults.

Future developments of this work will consider extending the application and assessment of the DropIn methodology to other neural models, both recurrent and static, also in conjunction with different case studies from those considered in this paper. On a longer term, instead, we are interested in studying if the same technique can be exploited to efficiently train recursive neural network models that can deal with structured data where different samples are characterized by different connectivity and topology \cite{DBLP:journals/tnn/BacciuMS12}.

\section*{Acknowledgment}
This work has been partially supported by the industrial research project funded by Biobeats Ltd a the Dipartimento di Informatica, Universit\`a di Pisa. D. Bacciu would like to acknowledge support from the Italian Ministry of Education, University, and Research (MIUR) under project SIR 2014 LIST-IT (grant n. RBSI14STDE).


\end{document}